\def\year{2018}\relax
\documentclass[letterpaper]{article} %DO NOT CHANGE THIS
\usepackage{aaai18}  %Required
\usepackage{times}  %Required
\usepackage{helvet}  %Required
\usepackage{courier}  %Required
\usepackage{url}  %Required
\usepackage{graphicx}  %Required
\usepackage{mathtools}
\usepackage{amsmath} % provides many mathematical environments & tools
\usepackage{amsfonts}
\usepackage{subcaption}
\usepackage{multirow}
\usepackage{stmaryrd}
\usepackage{algorithm}
\usepackage{algorithmic}
\newcommand{\cQ}{\mathcal{Q}}
\newcommand{\cR}{\mathcal{R}}
\newcommand{\fromto}{\longrightarrow}
\newcommand{\on}{\operatorname}
\DeclarePairedDelimiter{\indic}{\llbracket}{\rrbracket}
\newcommand{\argsort}{\operatorname*{arg\,sort}}

\title{Learning to Rank based on Analogical Reasoning}
\author{Mohsen Ahmadi Fahandar \and Eyke H{\"u}llermeier\\
	Department of Computer Science, Paderborn University \\
	Pohlweg 49-51, 33098 Paderborn, Germany\\
	ahmadim@mail.upb.de, eyke@upb.de\\
}
 
\begin{document}

%\author{Anonymous Author(s)\\
%Affiliation\\
%Address\\
%e-mail\\
%Paper ID: 3850 
%%Content areas: Preferences/Ranking Learning (primary),  Common-Sense Reasoning,\\ Case-Based Reasoning, Transfer, Adaptation, Multitask Learning Change
%}

\maketitle
\begin{abstract}
Object ranking or ``learning to rank'' is an important problem in the realm of preference learning. On the basis of training data in the form of a set of rankings
of objects represented as feature vectors, the goal is to learn a
ranking function that predicts a linear order of any new set of objects.
In this paper, we propose a new approach to object ranking based on principles
of analogical reasoning. More specifically, our inference pattern is formalized in terms of so-called analogical proportions and can be summarized as follows: Given objects $A,B,C,D$, if object $A$ is known to be preferred to $B$, and $C$ relates to $D$ as $A$ relates to $B$, then $C$ is (supposedly) preferred to $D$. Our method applies this pattern as a main building block and combines it with ideas and techniques from instance-based learning and rank aggregation. Based on first experimental results for data sets from various domains (sports, education, tourism, etc.), we conclude that our approach is highly competitive. It appears to be specifically interesting in situations in which the objects are coming from different subdomains, and which hence require a kind of knowledge transfer.
\end{abstract}

\section{Introduction}

Preference learning has received increasing attention in machine learning in recent years \cite{mpub218}. Roughly speaking, the goal in preference learning is to induce preference models from observational (or experimental) data that reveal information about the preferences of an individual or a group of individuals in a direct or indirect way; the latter typically serve the purpose of predictive modeling, i.e., they are then used to predict the preferences in a new situation. 

In general, a preference learning system is provided with a set of items (e.g., products) for which preferences are known, and the task is to learn a function that predicts preferences for a new set of items (e.g., new products not seen so far), or for the same set of items in a different situation (e.g., the same products but for a different user). Frequently, the predicted preference relation is required to form a total order, in which case we also speak of a \emph{ranking problem}. In fact, among the problems in the realm of preference learning, the task of ``learning to rank'' has probably received the most attention in the literature so far, and a number of different ranking problems have already been introduced. Based on the type of training data and the required predictions, \citeauthor{mpub218} distinguish between the problems of  object ranking \cite{Cohen99,kami_as10}, label ranking \cite{Har-Peled2002,Cheng2009,Vembu2011}, and instance ranking \cite{mpub191}.

\begin{figure}
\begin{center}
	\begin{subfigure}{.15\textwidth}
		\centering
		\includegraphics[height=1.5cm]{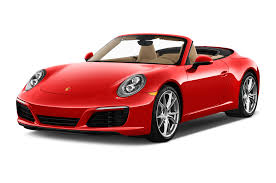}
		\caption*{$A$}
	\end{subfigure}
	\begin{subfigure}{.15\textwidth}
		\centering
		\includegraphics[height=1.5cm]{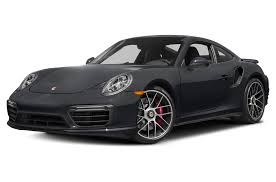}
		\caption*{$B$}
	\end{subfigure}
	
	\begin{subfigure}{.15\textwidth}
		\centering
		\includegraphics[height=1.5cm]{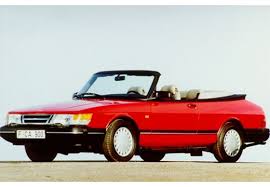}
		\caption*{$C$}
	\end{subfigure}
	\begin{subfigure}{.15\textwidth}
		\centering
		\includegraphics[height=1.5cm]{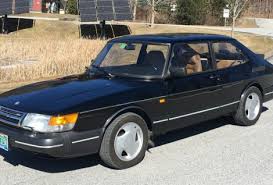}
		\caption*{$D$}
	\end{subfigure}
	\caption{$A$ is to $B$ as $C$ is to $D$ (pictures from ImageNet).}
	\label{fig:analogy}
	\end{center}
\end{figure}

The focus of this paper is on object ranking. Given training data in the form of a set of exemplary rankings of subsets of objects, the goal in object ranking is to learn a ranking function that is able to predict the ranking of any new set of objects.
Our main contribution is a novel approach that is based on the idea of \emph{analogical reasoning}, and essentially builds on the following inference pattern: If object $A$ relates to object $B$ as $C$ relates to $D$, and knowing that $A$ is preferred to $B$, we (hypothetically) infer that $C$ is preferred to $D$. Figure \ref{fig:analogy} provides an illustration with cars as objects. Cars $A$ and $B$ are more or less the same, except that $A$ is a cabriolet and has color red instead of black, and the same is true for cars $C$ and $D$. Thus, knowing that someone likes car $A$ more than $B$, we may conjecture that the same person prefers car $C$ to $D$. Our method applies this pattern as a main building block and combines it with ideas and techniques from instance-based learning and rank aggregation.

The rest of the paper is organized as follows. In the next section, we recall the setting of object ranking and formalize the corresponding learning problem. In the next section, we present a formalization of analogical reasoning based on the concept of analogical proportion. Next, we introduce our approach to \textbf{a}nalogy-\textbf{b}ased \textbf{le}arning to rank (able2rank). Finally, we present an experimental evaluation of this approach, prior to concluding the paper with a summary and an outline of future work.

\section{Learning to Rank}

Consider a reference set of objects, items, or choice alternatives $\mathcal{X}$, and assume each item $\boldsymbol{x} \in \mathcal{X}$ to be described in terms of a feature vector; thus, an item is a vector $\boldsymbol{x}  = (x_1, \ldots , x_d) \in \mathbb{R}^d$ and $\mathcal{X} \subseteq \mathbb{R}^d$. 
The goal in object ranking is to learn a \emph{ranking function} $\rho$ that accepts any (query) subset 
$$
Q = \{ \boldsymbol{x}_1, \ldots , \boldsymbol{x}_n \} \subseteq \mathcal{X}
$$
of $n = |Q|$ items as input. As output, the function produces a ranking $\pi \in \mathbb{S}_n$ of these items, where $\mathbb{S}_n$ denotes the set of all permutations of length $n$, i.e., all mappings $[n] \fromto [n]$ (symmetric group of order $n$); $\pi$ represents the total order
\begin{equation}\label{eq:r}
\boldsymbol{x}_{\pi^{-1}(1)} \succ \boldsymbol{x}_{\pi^{-1}(2)} \succ \ldots \succ \boldsymbol{x}_{\pi^{-1}(n)} \enspace ,
\end{equation}
i.e., $\pi^{-1}(k)$ is the index of the item on position $k$, while $\pi(k)$ is the position of the $k$th item $\boldsymbol{x}_k$ ($\pi$ is often called a \emph{ranking} and $\pi^{-1}$ an \emph{ordering}). Formally, a ranking function is thus a mapping
\begin{equation}\label{eq:map}
\rho: \, \cQ \fromto \cR \enspace ,
\end{equation}
where $\cQ = 2^\mathcal{X} \setminus \emptyset$ is the \emph{query space} and $\cR = \bigcup_{n \in \mathbb{N}} \mathbb{S}_n$ the \emph{ranking space}. The order relation $\succ$ is typically (though not necessarily) interpreted in terms of preferences, i.e., $\boldsymbol{x} \succ \boldsymbol{y}$ suggests that $\boldsymbol{x}$ is preferred to $\boldsymbol{y}$. 

\begin{figure}
\begin{center}
\includegraphics[width=0.45\textwidth]{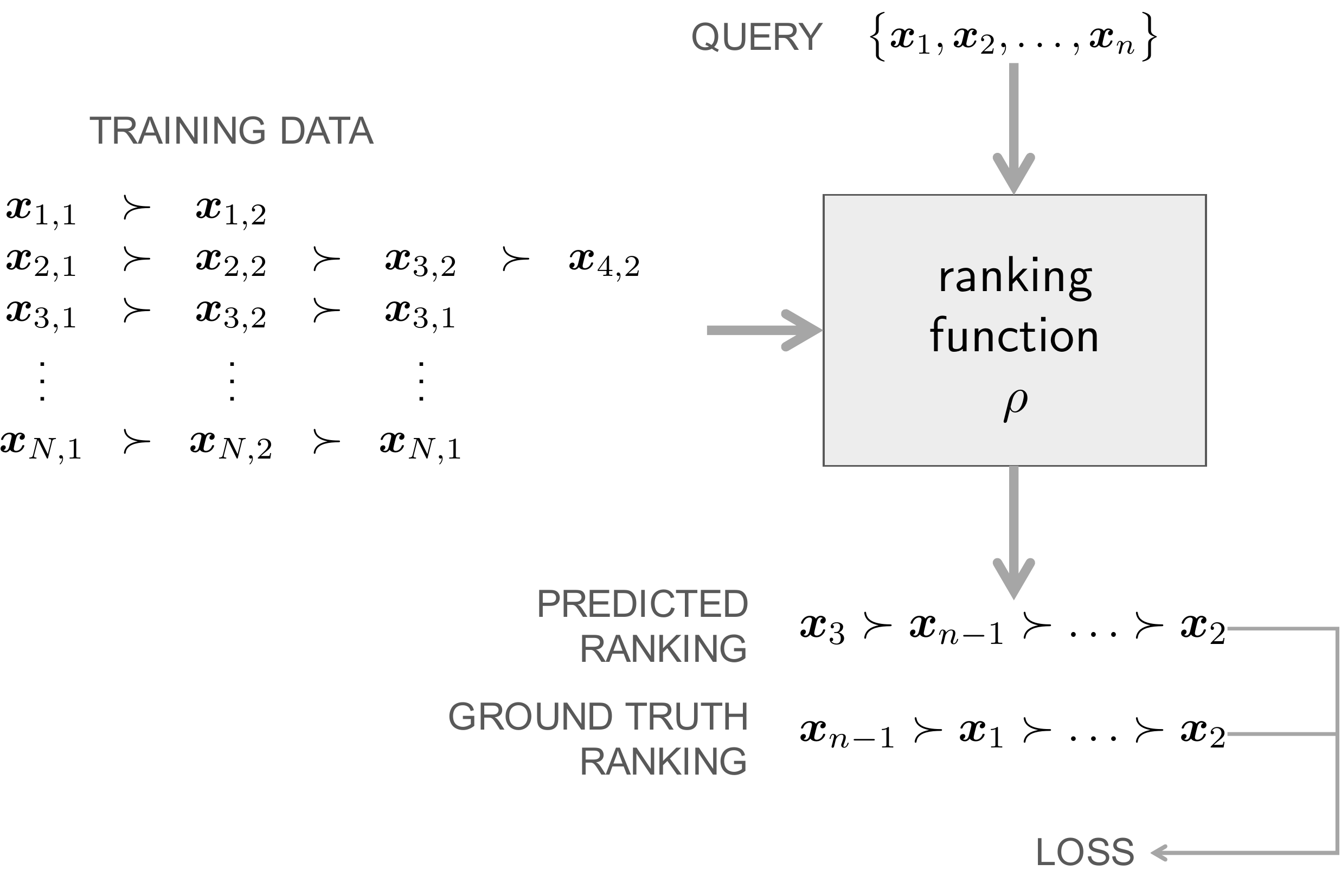}
\caption{The setting of object ranking.}
\label{fig:or}
\end{center}
\end{figure}

A ranking function $\rho$ is learned on a set of training data that consists of a set of rankings 
\begin{equation}\label{eq:td}
\mathcal{D} =  \big\{ (Q_1, \pi_1) , \ldots , (Q_M, \pi_M) \big\} \, , 
\end{equation}
where each ranking $\pi_j$ defines a total order of the query set $Q_j$. Once a ranking function has been learned, it can be used for making predictions for new query sets $Q$ (see Figure \ref{fig:or}). Such predictions are evaluated in terms of a suitable loss function or performance metric. A common choice is the (normalized) \emph{ranking loss}, which counts the number of inversions between two rankings $\pi$ and $\pi'$:
\begin{equation*} \label{eq:rankloss}
	d_{RL}(\pi, \pi') = 
		\frac{
	    \sum_{1 \leq i , j \leq n}  \indic{\pi(i) < \pi(j)} \indic{\pi'(i) > \pi'(j)}}{n(n-1)/2} \, ,
\end{equation*}
where $\llbracket \cdot \rrbracket$ is the indicator function.

\subsection{Methods}
The ranking function (\ref{eq:map}) sought in object ranking is a complex mapping from the query to the ranking space. An important question, therefore, is how to represent a ``ranking-valued'' function of that kind, and how it can be learned efficiently. As will be seen later on, our approach avoids an explicit representation of this function, and instead implements a query-specific (transductive) inference, in very much the same way as instance-based learning represents a predictor (through the sample data) in an indirect way.  

Most commonly, a ranking function is represented by means of an underlying scoring function 
$$
U:\, \mathcal{X} \fromto \mathbb{R} \, , 
$$
so that $\boldsymbol{x} \succ \boldsymbol{x}'$ if $U(\boldsymbol{x})> U(\boldsymbol{x}')$. In other words, a ranking-valued function is represented through a real-valued function. Obviously, $U$ can be considered as a kind of utility function, and $U(\boldsymbol{x})$ as a latent utility degree assigned to an item $\boldsymbol{x}$. Seen from this point of view, the goal in object ranking is to learn a latent utility function on a reference set $\mathcal{X}$.

The representation of a ranking function in terms of a real-valued (utility) function also suggests natural approaches to learning. In particular, two such approaches are prevailing in the literature. The first one reduces the original ranking problem to \emph{regression}; as it seeks a model that assigns appropriate scores to individual items $\boldsymbol{x}$, it is referred to as the \emph{pointwise} approach in the literature. The second idea is to reduce the problem to \emph{binary classification}; here, the focus is on pairs of items, which is why the approach is called the \emph{pairwise} approach.

As a representative of the first category, we will include \emph{expected rank regression} (ERR) in our experimental study later on \cite{Kamishima2006,kami_as10}. ERR reduces object ranking to standard linear regression. To this end, every training example $(Q, \pi)$ is replaced by a set of data points $(\boldsymbol{x}_i , y_i) \in \mathcal{X} \times \mathbb{R}$. Here, the target $y_i$ assigned to object $\boldsymbol{x}_i \in Q$ is given by 
$$
y_i = \frac{\pi(i)}{|Q| + 1} \ .
$$
This is justified by taking an expectation over all (complete) rankings of $\mathcal{X}$ and assuming a uniform distribution. In spite of this apparently oversimplified assumption, and the questionable transformation of ordinal ranks into numerical scores, ERR has shown quite competitive performance in empirical studies, especially when all rankings in the training data (\ref{eq:td}) are of approximately the same length \cite{tfml}. 
 
Given a ranking (\ref{eq:r}) as training information, the pairwise approach extracts all pairwise preferences $\boldsymbol{x}_{\pi^{-1}(i)} \succ \boldsymbol{x}_{\pi^{-1}(j)}$, $1 \leq i < j \leq n$, and considers these preferences as examples for a binary classification task. This approach is especially simple if $U$ is a linear function of the form $U(\boldsymbol{x}) = \boldsymbol{w}^\top \boldsymbol{x}$. In this case, $U(\boldsymbol{x}) > U(\boldsymbol{x}')$ if $\boldsymbol{w}^\top \boldsymbol{x} > \boldsymbol{w}^\top \boldsymbol{x}'$, which is equivalent to $\boldsymbol{w}^\top \boldsymbol{z} > 0$ for $\boldsymbol{z} = \boldsymbol{x} - \boldsymbol{x}' \in \mathbb{R}^d$. Thus, from the point of view of binary classification (with a linear threshold model), $\boldsymbol{z}$ can be considered as a positive and $-\boldsymbol{z}$ as a negative example. In principle, any binary classification algorithm can be applied to learn the weight vector $\boldsymbol{w}$ from the set of examples produced in this way. As a representative of this class of methods, we will use support vector machines in our experiments; more specifically, we include Ranking SVM \cite{joachims02} as a state-of-the-art baseline to compare with.

\section{Analogical Reasoning}

Analogical reasoning has a long tradition in artificial intelligence research, and various attempts at formalizing analogy-based inference can be found in the literature. We are building on a recent approach that is based on the concept of \emph{analogical proportion} \cite{Miclet2009,prade17}, which has already been used successfully in different problem domains, including classification \cite{bounhas14}, recommendation \cite{hug16}, preference completion \cite{pirlot16}, decision making \cite{billingsley17} and solving IQ tests \cite{beltran16}.

Consider four values $a, b, c, d$ from a domain $\mathbb{X}$. The quadruple $(a,b,c,d)$ is said to be in analogical proportion, denoted $a:b::c:d$, if ``$a$ relates to $b$ as $c$ relates to $d$''. A bit more formally, this idea can be expressed as
\begin{equation}\label{eq:ap}
\mathcal{R}(a,b) \sim \mathcal{R}(c,d) \, ,
\end{equation}
where the relation $\sim$ denotes the ``as'' part of the informal description. $\mathcal{R}$ can be instantiated in different ways, depending on the underlying domain $\mathbb{X}$. Specifically relevant for us are the cases of Boolean variables, where $\mathbb{X} = \mathbb{B} = \{0,1\}$, and numerical variables, where $\mathbb{X} = \mathbb{R}$. In the latter case, we will distinguish between \emph{arithmetic proportions}, where $\mathcal{R}(a,b) = a -b$, and \emph{geometric proportions}, where $\mathcal{R}(a,b) = a/b$.

Independently of the concrete choice of $\mathbb{X}$ and $\mathcal{R}$, the following axioms can reasonably be required (assuming an expression $a:b::c:d$ to be either true or false):
\begin{itemize}
	\item $a:b::a:b$ (reflexivity)
	\item $a:b::c:d \Rightarrow c:d::a:b$ (symmetry)
	\item $a:b::c:d \Rightarrow a:c::b:d$ (central permutation)
\end{itemize}

\subsection{Boolean Variables}

Consider the case of Boolean variables with values in $\mathbb{B}=\{0,1\}$. There are $2^4=16$ instantiations of the pattern $a:b::c:d$. The smallest (and hence most informative) subset of \emph{logical} proportions satisfying the above axioms consists of the following 6 instantiations:
\begin{center}
\begin{tabular}{cccc}
\hline
$a$ & $b$ & $c$ & $d$ \\
\hline
0 & 0 & 0 & 0 \\
0 & 0 & 1 & 1 \\
0 & 1 & 0 & 1 \\
1 & 0 & 1 & 0 \\
1 & 1 & 0 & 0 \\
1 & 1 & 1 & 1 
\end{tabular}
\end{center}
This formalization captures the idea that $a$ differs from $b$ (in the sense of being ``equally true'', ``more true'', or ``less true'') exactly as $c$ differs from $d$, and vice versa. This can also be expressed by the following formula:
\begin{equation}\label{eq:f1}
((a \Rightarrow  b) \equiv (c \Rightarrow  d))
\wedge
((b \Rightarrow  a) \equiv (d \Rightarrow  c))
\end{equation}
We remark that the above definition further satisfies the properties of independence with regard to positive or negative encoding of features ($a:b::c:d \Rightarrow \neg a: \neg b :: \neg c : \neg d$) as well as transitivity ($a:b::c:d$ and $c:d::e:f$ $\Rightarrow$ $a:b::e:f$). For an in-depth discussion of logical proportions, we refer to \cite{prad_fa13}.

\subsection{Real-Valued Variables}

To extend analogical proportions to real-valued variables $X$, we assume these variables take values in $[0,1]$. Practically, this may require a normalization of the original variable in a preprocessing step (a discussion of this will follow below). 

Values in the unit interval can be interpreted as generalized (graded) truth degrees. For instance, if $x$ is the speed of a car, the normalized value $x=0.8$ can be interpreted as the degree to which the property ``high speed'' applies, i.e., the truth degree of the proposition ``the car has high speed''. The idea, then, is to generalize the formalization of analogical proportions in the Boolean case using generalized logical operators \cite{dubois16}. Naturally, the analogical proportion itself will then become a matter of degree, i.e., a quadruple $(a,b,c,d)$ can be in analogical proportion \emph{to some degree}; in the following, we will denote this degree by $v(a,b,c,d)$. 

For example, taking (\ref{eq:f1}) as a point of departure, and generalizing the logical conjunction $x \wedge y$ by the minimum operator $\min(x,y)$, the implication $x \Rightarrow y$ by Lukasiewicz implication $\max(1-x+y, 0)$, and the equivalence $x \equiv y$  by $1- |x-y|$, one arrives at the following expression:
$$
v_A(a,b,c,d) = 
1- | (a-b) - (c-d)| 
$$
if $\on{sign}(a-b) = \on{sign}(c-d)$ and
$$
v_A(a,b,c,d) =  1- \max( |a-b|, |c-d|) 
$$
otherwise. We also consider a variant $v_{A'}$ of this expression, which is given by $v_{A'}(a,b,c,d)=v_A(a,b,c,d)$ if $\on{sign}(a-b) = \on{sign}(c-d)$ and $v_{A'}(a,b,c,d)=0$ otherwise.

Based on alternative interpretations of $\mathcal{R}$ and $\sim$ in (\ref{eq:ap}), other proposals for measuring the degree of analogical proportion can be found in the literature:

\begin{itemize}
	\item  Geometric proportion \cite{Beltran2014}:
	\[
	v_G(a,b,c,d) = \dfrac{\min(ad,bc)}{\max(ad,bc)}  \enspace ,
	\]
	if
	$
	\text{sign}(a-b)=\text{sign}(c-d)
	$ and $\max(ad,bc) > 0$, and 0 otherwise.
	
	\item Min-Max proportion  \cite{Bounhas2014}:
	\[
	\begin{split}
	v_{MM}(a,b,c,d) = 1-  & \max \big ( \,  | \min(a,d)-\min(b,c) |, \\
	 & \; | \max(a,d)-\max(b,c) | \, \big ).
	\end{split}
	\]
	
	\item Approximate equality proportion \cite{Beltran2014}:
	\begin{align*}
	v_{AE}(a,b,c,d) =   \max \big( \, 
	   & \llbracket a \approx b \rrbracket \llbracket c \approx d \rrbracket , \\
	  &  \llbracket a \approx c \rrbracket \llbracket b \approx d \rrbracket  
	  \, \big ) \enspace ,
	\end{align*}
	where $x \approx y$ is true if $ |x-y| \le \epsilon $ for a fixed threshold $\epsilon \in [0,1]$. We will also use a graded variant of $v_{AE}(\cdot)$, denoted as $v_{AE'}$, as defined above. To this end, we generalize the indicator function and evaluate the approximate equality of values $a,b$ as $\llbracket a \approx b \rrbracket = \max(1- |a-b|/\epsilon , 0)$.    

\end{itemize}

\subsection{Extension to Feature Vectors} 

The degree of analogical proportion defined above can be generalized from individual values to objects in the form of vectors $\boldsymbol{x}=(x_1, \ldots , x_d)$, where $x_i \in \mathbb{X}_i \in \{ \mathbb{B} , \mathbb{R} \}$---in the context of object ranking, $\boldsymbol{x}$ is a feature vector characterizing a choice alternative. To this end, the degrees of analogical proportion for the individual entries are aggregated into a single degree:
\begin{equation}\label{eq:agg}
v(\boldsymbol{a}, \boldsymbol{b} , \boldsymbol{c} , \boldsymbol{d}) = \on{AGG} \big\{ 
v(a_i , b_i , c_i , d_i) \, \vert \, i = 1, \ldots , d \big\}
\end{equation}
The aggregation operator $\on{AGG}$ can be specified in different ways. For example, a conjunctive combination is obtained with $\on{AGG} = \min$. This aggregation is very strict, however, and does not allow for compensating a low value of analogical proportion on a single feature by large values on other features. In our current implementation of analogy-based object ranking, we therefore use the arithmetic average as an aggregation.

\section{Analogy-based Object Ranking}

Recall that, in the setting of learning to rank, we suppose to be given a set of training data in the form 
$$
\mathcal{D} = \big\{ (Q_1, \pi_1) , \ldots , (Q_M, \pi_M) \big\} \ ,
$$
where each $\pi_m$ defines a ranking of the set of objects $Q_m$. If $\boldsymbol{z}_i , \boldsymbol{z}_j \in Q_m$ and $\pi_m(i) < \pi_m(j)$, then $\boldsymbol{z}_i \succ \boldsymbol{z}_j$ has been observed as a preference. In the following, we will denote by 
$$
\mathcal{D}_{pair} = \bigcup_{m=1}^M  \bigcup_{1 \leq i < j \leq |Q_m|}
(\boldsymbol{z}_{\pi^{-1}(i)}, \boldsymbol{z}_{\pi^{-1}(j)}) 
$$
the set of all pairwise preferences that can be extracted from $\mathcal{D}$.
The goal is to generalize beyond the data $\mathcal{D}$ so as to be able to predict a ranking of any query set 
$$
Q = \{ \boldsymbol{x}_1, \ldots , \boldsymbol{x}_n \} \, .
$$
Our analogy-based approach to object ranking (able2rank) consists of two main steps:
\begin{itemize}
\item First, for each pair of objects $\boldsymbol{x}_i , \boldsymbol{x}_j \in Q$, a degree of preference $p_{i,j} \in [0,1]$ is extracted from $\mathcal{D}$. If these degrees are normalized such that $p_{i,j} + p_{j,i} = 1$, they define a reciprocal preference relation
$$
P= \Big( p_{i,j} \Big)_{1 \leq i , j \leq n} \, .
$$ 
Normalization is not a strict requirement, however; in our concrete implementation, pairwise preference are expressed in terms of absolute frequencies (see below).
\item Second, the preference relation $P$ is turned into a ranking $\pi$ using a suitable ranking procedure. 
\end{itemize}
Both steps will be explained in more detail further below. Before, however, we briefly return to the issue of data preprocessing.

\subsection{Preprocessing}

Recall that we assume objects to be represented in terms of feature vectors $\boldsymbol{x}=(x_1, \ldots , x_d)$.
As already mentioned, analogical reasoning based on analogical proportions assumes real-valued features to be normalized, with values in the unit interval $[0,1]$. Therefore, if the $k$th feature is real-valued, we apply the linear transformation 
	\[
	x_k  \leftarrow \dfrac{x_k -\min_k}{ \max_k - \min_k } \,
	\]
where $\min_k$ and $\max_k$ denote, respectively, the smallest and largest value of that feature in the data. This transformation is applied separately to the training and the test data. 

Prior to normalization, we apply a logarithmic transformation to some of the features. For each feature, we compute a numeric measure of the skewness of its distribution and the distribution of its log-transform \cite{skewness98}. If the skewness of the latter is smaller than the skewness of the former, the log-transform is adopted, i.e., each value $x_k$ is replaced by $\log(x_k)$; otherwise, the feature is left unchanged.

\subsection{Analogical Prediction of Pairwise Preferences}

The first step of able2rank, prediction of pairwise preferences, is based on analogical reasoning. The basic idea is as follows: Consider any pair of query objects $\boldsymbol{x}_i , \boldsymbol{x}_j \in Q$. Moreover, suppose we find a pair of objects $(\boldsymbol{z}, \boldsymbol{z}') \in \mathcal{D}_{pair}$ in the training data, i.e., a preference $\boldsymbol{z} \succ \boldsymbol{z}'$, such that $(\boldsymbol{z}, \boldsymbol{z}', \boldsymbol{x}_i , \boldsymbol{x}_j)$ are in analogical proportion. Then, this is taken as an indication in favor of the preference $\boldsymbol{x}_i \succ \boldsymbol{x}_j$. We refer to this principle as \emph{analogical transfer} of preferences, because the observed preference $\boldsymbol{z} \succ \boldsymbol{z}'$ between objects $\boldsymbol{z}, \boldsymbol{z}'$ is (hypothetically) transferred to $\boldsymbol{x}_i, \boldsymbol{x}_j$.

Since the training data is not necessarily coherent, both preferences $\boldsymbol{x}_i \succ \boldsymbol{x}_j$ and $\boldsymbol{x}_j \succ \boldsymbol{x}_i$ might be supported (via analogical transfer). We define 
$$
p_{i,j}=  \frac{c_{i,j}}{c_{i,j}+c_{j,i}} \, ,
$$
where $c_{i,j}$ is the number of preferences that support $\boldsymbol{x}_i \succ \boldsymbol{x}_j$, and $c_{j,i}$ the number of preferences supporting $\boldsymbol{x}_j \succ \boldsymbol{x}_i$. Here, instead of counting all preferences, we only consider the most relevant ones, i.e., those with the largest degrees of analogical proportion. More specifically, let 
$V = V_{i,j} \cup V_{j,i}$ with 
\begin{align}\label{eq:vij}
V_{i,j} & = \bigcup_{(\boldsymbol{z}, \boldsymbol{z}') \in \mathcal{D}_{pair}} v(\boldsymbol{z},\boldsymbol{z}',\boldsymbol{x}_i, \boldsymbol{x}_j)\\ V_{j,i} & = \bigcup_{(\boldsymbol{z}, \boldsymbol{z}') \in \mathcal{D}_{pair}} v(\boldsymbol{z},\boldsymbol{z}',\boldsymbol{x}_j, \boldsymbol{x}_i)
\label{eq:vji}
\end{align}
be the set of analogy scores for $\boldsymbol{x}_i$ and $\boldsymbol{x}_j$, and $V^* \subset V$ the $k$ largest scores in $V$. Then 
\begin{equation}\label{eq:cij}
c_{i,j} = | V_{i,j} \cap V^*| , \quad  c_{j,i} = | V_{j,i} \cap V^*| \, .
\end{equation}
The degrees of analogical proportion in (\ref{eq:vij}--\ref{eq:vji}) are given by (\ref{eq:agg}) with $v \in \{ v_A, v_{A'}, v_G, v_{MM}, v_{AE}, v_{AE'} \}$. Thus, $v$ plays the role of a parameter of our method, just like the value $k$ of relevant proportions taken into account; in our current implementation, we choose $k \in \{10, 15, 20 \}$. 
Algorithm (\ref{alg:AOR}) provides a summary of this part of the method.
\begin{algorithm}
	\caption{Analogy-based Pairwise Preferences (APP)}
	\label{alg:AOR}
	\begin{algorithmic}[1]
		\REQUIRE $\mathcal{D}_{pair}, Q$
		\STATE $lstX_iX_j \leftarrow []$  \emph{\small // initialize the list}
		\FORALL{$\boldsymbol{x}_i,\boldsymbol{x}_j \in Q$}
		\STATE \emph{\small// binary vector: supporting $\boldsymbol{x}_i \succ \boldsymbol{x}_j$ as 1 and 0 otherwise}
		\STATE $X_iX_j \leftarrow []$ 
		\STATE $scores \leftarrow []$ \emph{\small // numeric vector: the degree of support}
		\FORALL{$\boldsymbol{z}, \boldsymbol{z}^{\prime} \in \mathcal{D}_{pair}$}
		\STATE $s_{ij} \leftarrow 
		v(\boldsymbol{z},
		\boldsymbol{z}^{\prime},
		\boldsymbol{x}_i,
		\boldsymbol{x}_j)$
		\STATE $s_{ji} \leftarrow v(\boldsymbol{z},
		\boldsymbol{z}^{\prime},
		\boldsymbol{x}_j,
		\boldsymbol{x}_i)$
		\IF{$s_{ij}>s_{ji}$} 
		\STATE $scores.append(s_{ij})$ 
		\ELSE 
		\STATE $scores.append(s_{ji})$
		\ENDIF
		\STATE $X_iX_j.append( \; \text{not} (\text{xor} \big (\llbracket \boldsymbol{z} \succ \boldsymbol{z}^{\prime} \rrbracket, \llbracket s_{ij}>s_{ji} \rrbracket \big ))  \; )$
		\ENDFOR
		\STATE \emph{\small // re-arrange $X_iX_j$ based on positions of sorted $scores$}
		\STATE $X_iX_j \leftarrow X_iX_j( \; \argsort(scores) \; )$
		\STATE $lstX_iX_j.append(X_iX_j)$
		\ENDFOR
		\RETURN $lstX_iX_j$
	\end{algorithmic}
\end{algorithm}
\subsection{Rank Aggregation}

To turn pairwise preferences into a total order, we make use of a rank aggregation method. More specifically, we apply the Bradley-Terry-Luce (BTL) model, which is well-known in the literature on discrete choice \cite{brad_tr52}. It starts from the parametric model 
\begin{equation}\label{eq:pmp}
\mathbf{P}(\boldsymbol{x}_i \succ \boldsymbol{x}_j) = \frac{\theta_i}{\theta_i + \theta_j} \, ,
\end{equation}
where $\theta_i, \theta_j \in \mathbb{R}_+$ are parameters representing the (latent) utility $U(\boldsymbol{x}_i)$ and $U(\boldsymbol{x}_j)$ of $\boldsymbol{x}_i$ an $\boldsymbol{x}_j$, respectively. Thus, according to the BTL model, the probability to observe a preference in favor of a choice alternative $\boldsymbol{x}_i$, when being compared to any other alternative, is proportional to $\theta_i$. 

Given the data (\ref{eq:cij}), i.e., the numbers $c_{i,j}$ informing about how often every preference $\boldsymbol{x}_i \succ \boldsymbol{x}_j$ has been observed, the parameter $\theta = (\theta_1, \ldots , \theta_n)$ can be estimated by likelihood maximization:
$$
\hat{\theta} \in \arg \max_{\theta \in \mathbb{R}^{n} } \prod_{1 \leq i \neq j \leq n}  \left( \dfrac{\theta_{i}}{\theta_{i} + \theta_{j}} \right)^{c_{i,j}}
$$
Finally, the predicted ranking $\pi$ is obtained by sorting the items $\boldsymbol{x}_i$ in descending order of their estimated (latent) utilities $\hat{\theta}_i$ (see Algorithm \ref{alg:aggregation}). 

We note that many other rank aggregation techniques have been proposed in the literature and could principally be used as well; see e.g.\ \cite{pmlr-v70-fahandar17a}. However, since BTL seems to perform very well, we did not consider any other method.

\begin{algorithm}
	\caption{Rank Aggregation (RA)}
	\label{alg:aggregation}
	\begin{algorithmic}[1]
		\REQUIRE $lstX_iX_j, k$
		\STATE Initialize comparison matrix $C_{n \times n}$
		\FORALL{$X_iX_j$ in $lstX_iX_j$}
		\STATE $C_{i,j} \leftarrow \sum_{r=1}^{k} \llbracket X_iX_j(r)=1 \rrbracket$
		\STATE $C_{j,i} \leftarrow \sum_{r=1}^{k} \llbracket X_iX_j(r)=0 \rrbracket$
		\ENDFOR
		\STATE $\hat{\theta} \leftarrow BTL(C)$
		\STATE $\hat{\pi} \leftarrow \argsort(\hat{\theta})$
		\RETURN $\hat{\pi}$
	\end{algorithmic}
\end{algorithm}

\section{Experiments}
In order to study the practical performance of our proposed method, we conducted experiments on several real-world data sets, using ERR and SVM-Rank (cf.\ Section 2.1) as baselines for comparison.

\subsection{Data}

The data sets\footnote{available at \url{https://cs.uni-paderborn.de/is/}} are collected from various domains (e.g., sports, education, tourism) and comprise different types of feature (e.g., numeric, binary, ordinal). Table \ref{tab:ds} provides a summary of the characteristics of the data sets. Here is a detailed description:

\begin{table*}
	\caption{Properties of data sets.}
	\label{tab:ds}
	\small
	\centering
	\begin{tabular}{ |clcccccc| } 
		\hline
		data set & domain & \# instances & \# features & numeric & binary & ordinal & name \\
		\hline
		\multirow{4}{*}{Decathlon} & Year 2005 & 100 & 10 & x& -- & -- & D1\\ 
		& Year 2006 & 100 & 10 & x & -- & -- & D2 \\ 
		& Olympic games Rio de Janeiro 2016 & 24 & 10 & x & -- & -- & D3 \\ 
		& Under-20 world championships 2016 & 22 & 10 & x & -- & -- & D4 \\ 
		\hline
		\multirow{4}{*}{Bundesliga} & Season 15/16 & 18 & 13 & x& -- & -- & B1\\ 
		& Season 16/17 & 18 & 13 & x & -- & -- & B2 \\ 
		& Mid-Season 16/17 & 18 & 7 & x & -- & -- & B3 \\ 
		& Season 16/17 Away & 18 & 6 & x & -- & -- & B4 \\ 
		\hline
		%\multirow{2}{*}{FIFA} & Year 2016 & 100 & 40 & x& x & x & F1\\ 
		%& Year 2017 & 100 & 40 & x & x & x & F2 \\ 
		\multirow{4}{*}{FIFA} & Year 2016 & 100 & 40 & 36 & 1 & 3 & F1\\ 
		& Year 2017 & 100 & 40 & 36 & 1 & 3 & F2 \\ 
		& Year 2016 (Position:Streaker) & 50 & 40 & 36 & 1 & 3 & F3 \\ 
		& Year 2017 (Position:Streaker)& 50 & 40 & 36 & 1 & 3 & F4 \\ 
		\hline
		\multirow{2}{*}{Hotels} & D{\"u}sseldorf & 110 & 28 & x& x & x & H1\\ 
		& Frankfurt & 149 & 28 & x & x & x & H2 \\ 
		\hline
		\multirow{2}{*}{Uni. Rankings} & Year 2015 & 100 & 9 & x& -- & -- & U1\\ 
		& Year 2014 & 100 & 9 & x & -- & -- & U2 \\ 
		\hline
		\multirow{2}{*}{Volleyball WL} & Group 3 & 12 & 15 & x& -- & -- & V1\\ 
		& Group 1 & 12 & 15 & x & -- & -- & V2 \\ 
		\hline
		\multirow{2}{*}{Netflix} & Germany & 9 & 7 & x& x & -- & N1\\ 
		& USA & 13 & 7 & x & x & -- & N2 \\ 
		\hline
	\end{tabular}
\end{table*}

\begin{itemize}
	\item Decathlon: This data contains rankings of the top 100 men's decathletes worldwide in the years 2005 and 2006, with 10 numeric features associated with each athlete. Each feature is the performance achieved by the athlete in the corresponding discipline (e.g., the time in 100 meter race). The ranking of the decathletes is based on a scoring scheme, in which each performance is first scored in terms of a certain number of points (the mapping from performance to scores is non-linear), and the total score is obtained as the sum of the points over all 10 disciplines. In addition, the results of Olympic games Rio de Janeiro 2016 (24 instances) as well as under-20 world championships 2016 (22 instances) are considered. The data are extracted from the Decathlon2000 web site\footnote{\url{www.decathlon2000.com}}.
	
	\item FIFA: The FIFA Index website\footnote{\url{www.fifaindex.com}} ranks the best football players in the world based on different metrics each year. These metrics belong to different categories, such as ball skills (ball control, dribbling, etc.), physical performance (acceleration, balance, etc.), defence (marking, slide tackling, etc.), and so on. We considered the list of top 100 footballers in the years 2016 and 2017, where each player is described by 40 attributes. Since the metrics of different types of players are not comparable, the overall evaluation of a player depends on his position (goal keeper, defender, streaker, etc.). Obviously, predicting the overall ranking is therefore a difficult task. In addition to the full data sets, we therefore also considered two position-specific data sets, namely the rankings for players with position streaker in the years 2016 and 2017.
	
	\item Hotels: This data set contains rankings of hotels in two major German cities, namely D{\"u}sseldorf (110 hotels) and Frankfurt (149 hotels). These rankings have been collected from TripAdvisor\footnote{\url{www.tripadvisor.com}} in September 2014. Each hotel is described in terms of a feature vector of length 28 (e.g., distance to city center, number of stars, number of rooms, number of user rating, etc). The way in which a ranking is determined by TripAdvisor on the basis of these features is not known (and one cannot exclude that additional features may play a role).
	
	\item University Rankings: This data includes the list of top 100 universities worldwide for the years 2014 and 2015. It is published by the Center for World University Rankings (CWUR).  Each university is represented by 9 features such as national rank, quality of education, alumni employment, etc. Detailed information about how the ranking is determined based on these features can be found on the CWUR website\footnote{\url{www.cwur.org}}.
	
	\item Bundesliga: This data set comprises table standings of 18 football teams in the German Bundesliga (German football league) for the seasons 2015/16 and 2016/17.\footnote{\url{www.bundesliga.com}} Each team is described in terms of 13 features, such as matches, win, loss, draw, goals-for, goals-against, etc. To study the ability of knowledge transfer, we also included the table standing for the season 2016/17, in which only the statistics for away matches are considered (with 6 features). Another case is the table standing in the mid-season 2016/17 (i.e., only the first half of the season) with 7 features.
	
	\item Volleyball WL: This data contains the table standing for Group1 (statistics of 12 teams divided into subgroups, each with 9 matches) and Group3 (statistics of 12 teams, each with 6 matches) of volleyball world league 2017 extracted from the FIVB website\footnote{\url{worldleague.2017.fivb.com}}. There are 12 features in total, such as win, loss, number of (3-0, 3-1, 3-2, etc.) wins, sets win, sets loss, etc.
	
	\item Netflix: This data set includes the Netflix ISP speed index (extracted from Netflix website\footnote{\url{ispspeedindex.netflix.com}} in August, 2017) for Germany and USA with 9 and 13 Internet providers, respectively. The Netflix ISP Speed Index is a measure of prime time Netflix performance on particular internet service providers (ISPs) around the globe. The rankings are represented by 7 (binary and numeric) features like speed, speed of previous month, fiber, cable, etc.
\end{itemize}

\subsection{Experimental Setting}
Given the data sets for training and testing, able2rank first preprocesses the individual attributes as explained above (log-transformation and normalization). We also apply a standard normalization for the baseline methods (ERR and SVM), transforming each real-valued feature by standardization:
\[
x  \leftarrow \dfrac{x-\mu}{\sigma} \, ,
\]
where $\mu$ and $\sigma$ denote the empirical mean and standard deviation, respectively. 

Recall that able2rank has two parameters to be tuned: The type of analogical proportion $v \in \mathcal{S}_v$, where $\mathcal{S}_v = \{ v_A, v_{A'}, v_G, v_{MM}, v_{AE}, v_{AE'} \}$, and the number $k \in \mathcal{S}_k$ of relevant proportions considered for estimating pairwise preferences, where $\mathcal{S}_k = \{10,15,20\}$. We fixed these parameters in an (internal) 2-fold cross validation (repeated 5 times) on the training data, using simple grid search on $\mathcal{S}_v \times \mathcal{S}_k$ (i.e., trying all combinations).
The combination $(v^* , k^*)$ with the lowest cross-validated error $\on{d}_{\text{RL}}$ is eventually adopted and used to make predictions on the test data (using the entire training data).
The complexity parameter $C$ of SVM is fixed in a similar way using an internal cross-validation.

\subsection{Results}

In our experiments, predictions were produced for certain parts $\mathcal{D}_{test}$ of the data we collected, using other parts $\mathcal{D}_{train}$ as training data; an experiment of that kind is denoted $\mathcal{D}_{train} \rightarrow \mathcal{D}_{test}$. The results of the conducted experiments are summarized in Table \ref{tab:results}, with the best performance on each problem displayed in bold-face. Additionally, we report the parameters used by able2rank.

From the results, we conclude that able2rank is quite competitive in terms of predictive accuracy. In 8 of 11 cases, it achieves the best performance in terms of the average ranking loss $d_{RL}$ (which translates to a p-value of around 10\% when comparing able2rank and SVM with a pairwise sign test, and of around 0.5\% for the comparison with ERR).

\section{Conclusion and Future Work}

This paper advocates the use of analogical reasoning in the context of preference learning. Building on the notion of analogical proportion, we formalize the heuristic principle suggesting that, if an alternative $A$ is preferred to $B$, and $C$ relates to $D$ as $A$ relates to $B$, then $C$ is likely to be preferred to $D$. Based on this formalization, we develop a concrete method, able2rank, for the problem of object ranking. First experimental results on real-world data from different domains are quite promising and suggest that able2rank is competitive to state-of-the-art methods for object ranking.

\begin{table}
	\caption{Results in terms of loss $d_{RL}$ on the test data.}
	\label{tab:results}
	\begin{tabular}{ccccc}
\hline
train $\rightarrow$ test & $(v^*, k^*)$ & able2rank & ERR & SVM \\ 
\hline
D1 $\rightarrow$ D2 & $v_{A'},10$ & $0.066$ & $0.064$ & $\boldsymbol{0.025}$   \\ 
D3 $\rightarrow$ D4 & $v_{MM},10$ & $\boldsymbol{0.139}$ &  $0.147$ & $0.152$   \\ 
B1 $\rightarrow$ B2 & $v_{A},15$ & $\boldsymbol{0.046}$ &	$0.144$ & $0.059$        \\ 
B4 $\rightarrow$ B2 & $v_{A},20$ & $0.078$ &	$0.078$ & $\boldsymbol{0.026}$   \\ 
B3 $\rightarrow$ B2  & $v_{A},10$ & $\boldsymbol{0.000}$ &	$0.033$ & $0.013$  \\ 
F1 $\rightarrow$ F2 & $v_{G},20$ & $\boldsymbol{0.225}$ & $0.259$ & $0.310$ \\ 
F3 $\rightarrow$ F4 & $v_{G},10$  & $\boldsymbol{0.158}$ & $0.364$ & $0.169$\\ 
H1 $\rightarrow$ H2 & $v_{A},20$ & $\boldsymbol{0.060}$ & $0.084$ & $0.072$  \\ 
U1 $\rightarrow$ U2 & $v_{A'},15$ & $\boldsymbol{0.073}$ &	$0.163$ & $0.154$ \\ 
V1 $\rightarrow$ V2 & $v_{MM},10$ & $0.030$ & $0.485$ & $\boldsymbol{0.015}$ \\ 
N1 $\rightarrow$ N2  & $v_{A},10$ & $\boldsymbol{0.013}$ & $0.090$ & $0.077$ \\ 
\end{tabular}
\end{table}

In future work, we plan to elaborate more closely on the setting of \emph{transfer learning}, because we believe our analogical approach to preference learning, like analogical reasoning in general, to be specifically useful for this purpose. In fact, preference transfer as defined in this paper only requires the relation $\mathcal{R}$ to be evaluated separately for \emph{source} objects $A$ and $B$ on the one side and \emph{target} objects $C$ and $D$ on the other side, but never between sources and targets; in principle, different specifications of $\mathcal{R}$ could even be used for the source and the target. In the experiments conducted in this paper, some kind of knowledge transfer was already required, but source and target were still closely related and essentially from the same domain. As already said, we plan to extend these experiments toward problems where source and target are from different domains, and to generalize able2rank correspondingly.  

Besides, the basic version of able2rank as presented in this paper ought to be improved and extended in different directions. For example, for the time being, the analogical proportion of feature vectors in able2rank is simply the average of the analogical proportions on the individual attributes. However, since different features may have a different influence on preferences and analogies, the selection or weighting of features appears to be quite important. Therefore, the development of methods for feature selection and weighting will be addressed in future work.

Furthermore, the current version of able2rank ignores possible interactions between features. For example, while a person may prefer color red to black for T-shirts, this preference might be reversed in the case of pullovers. Learning such interactions and capturing them in analogical inference is a highly non-trivial problem.  

Last but not least, issues related to computational efficiency ought to be addressed. Since able2rank follows the lazy learning paradigm \cite{aha_ll}, just like instance-based learning and nearest neighbor prediction, it can learn very easily (simply by storing observed preferences) but is relatively costly at prediction time. In comparison to nearest neighbor classification, for example, the cost is even higher due to the need to iterate over \emph{pairs} of objects in the training data to find the highest analogical proportions. Thus, implementing ``highest analogy'' search in a naive way, the complexity scales quadratically with the size of the training data. Consequently, there is a need to reduce this complexity by means of suitable data structures and algorithms for efficient retrieval of analogy pairs. Besides, other means for complexity reduction could be considered, such as instance selection or editing strategies like those used in case-based reasoning \cite{mcke_cg00,dela_aa04}. 

%\section{ Acknowledgments}

\bibliography{AhmadiFahandar-Huellermeier_bibfile}
\bibliographystyle{aaai}

\end{document}